\documentclass{article} 
\usepackage{iclr2025_conference,times}

\usepackage{amsmath,amsfonts,bm}

\def\eqref#1{equation~\ref{#1}}

\def\1{\bm{1}}

\DeclareMathAlphabet{\mathsfit}{\encodingdefault}{\sfdefault}{m}{sl}
\SetMathAlphabet{\mathsfit}{bold}{\encodingdefault}{\sfdefault}{bx}{n}

\usepackage[utf8]{inputenc} 
\usepackage[T1]{fontenc}    
\usepackage{hyperref}       
\usepackage{url}            
\usepackage{booktabs}       
\usepackage{amsfonts}       
\usepackage{nicefrac}       
\usepackage{microtype}      
\usepackage{xcolor}         
\usepackage{xspace}
\usepackage{graphicx}
\usepackage{times}
\usepackage{latexsym}
\usepackage{multirow}
\usepackage{balance}
\usepackage{bbm}
\usepackage{paralist}
\usepackage{makecell}
\usepackage{wrapfig}
\usepackage{tabularx} 
\usepackage{amsmath}
\usepackage{amssymb}
\usepackage{mathtools}
\usepackage{amsthm}
\usepackage{multicol}
\usepackage{subcaption}
\usepackage{xcolor}
\usepackage{colortbl}
\usepackage{enumitem}
\usepackage{cleveref}

\usepackage{siunitx}
\usepackage{graphicx}

\usepackage{wrapfig}
\usepackage{enumitem}
\setlist[itemize]{left=3pt, before=\vspace{-2pt}} %
\usepackage{multirow}
\usepackage{xspace}
\usepackage{adjustbox}
\usepackage{pifont}
\usepackage{caption}
\usepackage{makecell}
\usepackage{nicematrix}
\usepackage{listings}
\usepackage{fontawesome}

\usepackage{array}
\usepackage{xtab}
\usepackage{arydshln}

\definecolor{darkred}{RGB}{156, 39, 33}
\definecolor{darkblue}{RGB}{31, 90, 153}
\definecolor{mylightblue}{rgb}{0.8, 0.9, 1.0}

\newcolumntype{L}[1]{>{\raggedright\let\newline\\\arraybackslash\hspace{0pt}}m{#1}}
\newcolumntype{C}[1]{>{\centering\let\newline\\\arraybackslash\hspace{0pt}}m{#1}}
\newcolumntype{R}[1]{>{\raggedleft\let\newline\\\arraybackslash\hspace{0pt}}m{#1}}
\newcolumntype{P}[1]{>{\centering\let\newline\\\arraybackslash\columncolor{mylightblue}}m{#1}}

\hypersetup{
    colorlinks,
    linkcolor={red!50!black},
    citecolor={blue!50!black},
    urlcolor={blue!80!black}
}

\title{FuseChat-3.0: Preference Optimization Meets Heterogeneous Model Fusion}

\author{Ziyi Yang, Fanqi Wan, Longguang Zhong, Canbin Huang, Guosheng Liang, Xiaojun Quan\thanks{$\;\;$Corresponding author.} \\
School of Computer Science and Engineering, Sun Yat-sen University, China \\
\texttt{yangzy39@mail2.sysu.edu.cn, quanxj3@mail.sysu.edu.cn} 
}

\iclrfinalcopy 
\begin{document}

\maketitle

\begin{abstract}
We introduce FuseChat-3.0, a suite of large language models (LLMs) developed by integrating the strengths of heterogeneous source LLMs into more compact target LLMs. Our source models include the powerful Gemma-2-27B-it, Mistral-Large-Instruct-2407, Qwen-2.5-72B-Instruct, and Llama-3.1-70B-Instruct. For target models, we focus on three widely-used smaller variants—Llama-3.1-8B-Instruct, Gemma-2-9B-it, and Qwen-2.5-7B-Instruct—along with two ultra-compact options, Llama-3.2-3B-Instruct and Llama-3.2-1B-Instruct. To leverage the diverse capabilities of these source models, we develop a specialized data construction protocol tailored to various tasks and domains. The FuseChat-3.0 training pipeline consists of two key stages: (1) supervised fine-tuning (SFT) to align the target and source model distributions, and (2) Direct Preference Optimization (DPO) to apply preferences from multiple source LLMs to fine-tune the target model. 
The resulting FuseChat-3.0 models exhibit significant performance gains across tasks such as \textit{instruction following}, \textit{general knowledge}, \textit{mathematics}, and \textit{coding}. As illustrated in Figure~\ref{fig:fusechat_llama3_8b}, using Llama-3.1-8B-Instruct as the target model, our fusion approach achieves an average improvement of 6.8 points across 14 benchmarks. Moreover, it demonstrates remarkable gains of 37.1 points and 30.1 points on the instruction-following benchmarks AlpacaEval-2 and Arena-Hard, respectively. Our code, models, and datasets are available at \url{https://github.com/SLIT-AI/FuseChat-3.0}.
\vspace{-0.22cm}
\begin{figure*}[ht]
    \centering
    \includegraphics[width=0.52\textwidth]{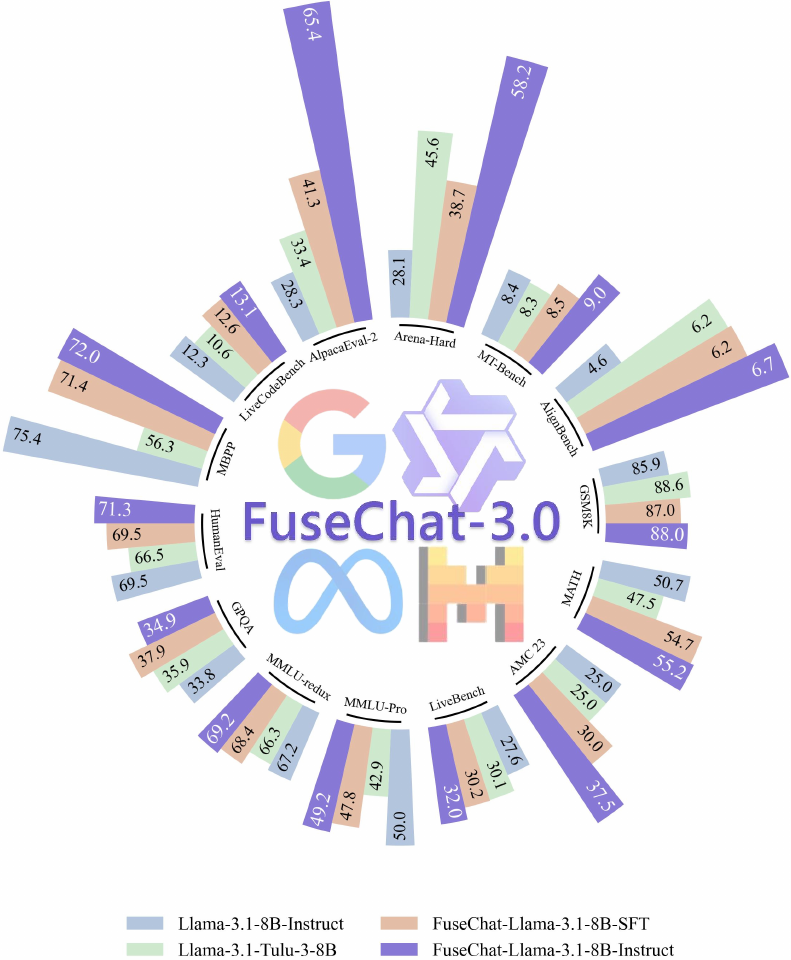}
    \vspace{-0.22cm}
 \caption{Results of FuseChat-3.0 across various benchmarks, using Llama-3.1-8B-Instruct as the target LLM.}
\label{fig:fusechat_llama3_8b}
\end{figure*}
\end{abstract}

\section{Introduction}
\vspace{-0.15cm}
\label{intro}
Combining the strengths of multiple large language models (LLMs) provides a powerful means to enhance performance, robustness, and generalization across diverse tasks by leveraging the unique expertise and knowledge each model offers. Individual LLMs, particularly those constrained by size or training data, may perform well in specific areas but struggle in others due to specialization gaps. For instance, one model might excel at generating creative content but lack precision in technical explanations, while another delivers technical accuracy but struggles with conversational fluency. By integrating multiple models, their collective strengths can bridge these gaps, leading to improved overall performance. This collaborative approach also improves robustness, as the system can compensate for individual model errors—when one model underperforms, others can intervene to support the response. Furthermore, this integration enhances task generalization by exposing the system to diverse patterns, allowing it to adapt more effectively to new or unseen challenges. 

Various strategies have been developed to achieve this, each with unique trade-offs. Ensemble methods~\citep{jiang2023llm, wang2024mixture} enhance performance and robustness by combining predictions from multiple models. However, they require all models to remain active during inference, leading to substantial computational and memory costs. LLM routing~\citep{ding2024hybrid,hu2024routerbench,ong2025routellm} offers a more efficient alternative: a router selects the most appropriate LLM to handle each query. While this balances effectiveness and efficiency, it requires training a new router for each task, limiting its generalization to unseen tasks. 
Model merging~\citep{wortsman2022model} integrates models with identical architectures into a unified parameter set, improving robustness and generalization but limiting applicability to homogeneous model families.
Explicit model fusion (EMF) methods~\citep{wan2024knowledge,wan2024fusechatknowledgefusionchat} use knowledge distillation to transfer knowledge from multiple source models to a single target model, often through probabilistic distribution matrices. While adaptable to different model structures and sizes, EMF faces challenges like vocabulary alignment and distribution merging, which can complicate the fusion process and introduce errors.

In this paper, we present FuseChat-3.0, a suite of large language models (LLMs) developed by harnessing the strengths of heterogeneous source LLMs into more compact target models. Our approach enables target LLMs to learn directly from the generated outputs of diverse open-source LLMs—a process termed implicit model fusion (IMF) \citep{yang2025weightedreward}. IMF leverages the rich information embedded in outputs from stronger LLMs through various mechanisms. For example, these outputs can be used to fine-tune weaker models~\citep{tian2024tinyllm,kang2024knowledge}. In the domain of preference learning, methods like Zephyr~\citep{tunstall2024zephyr} and T\"ulu 3~\citep{lambert2024t} utilize GPT-4 to rank responses generated by multiple LLMs, thereby creating preference datasets for Direct Preference Optimization (DPO) training~\citep{Rafailov2023DirectPO}. Similarly, WRPO constructs preference pairs by combining responses from both target and source models.
However, generating positive and negative examples from different models can introduce reward annotation bias and variance, potentially undermining optimization. In contrast, our method addresses these issues by constructing each preference pair from the best and worst responses generated by the same source model. This crucial distinction eliminates the reward bias caused by heterogeneous response styles, prevents reward hacking, and delivers more controlled preference signals.

\begin{figure*}[t]
    \centering
    \includegraphics[width=0.99\textwidth]{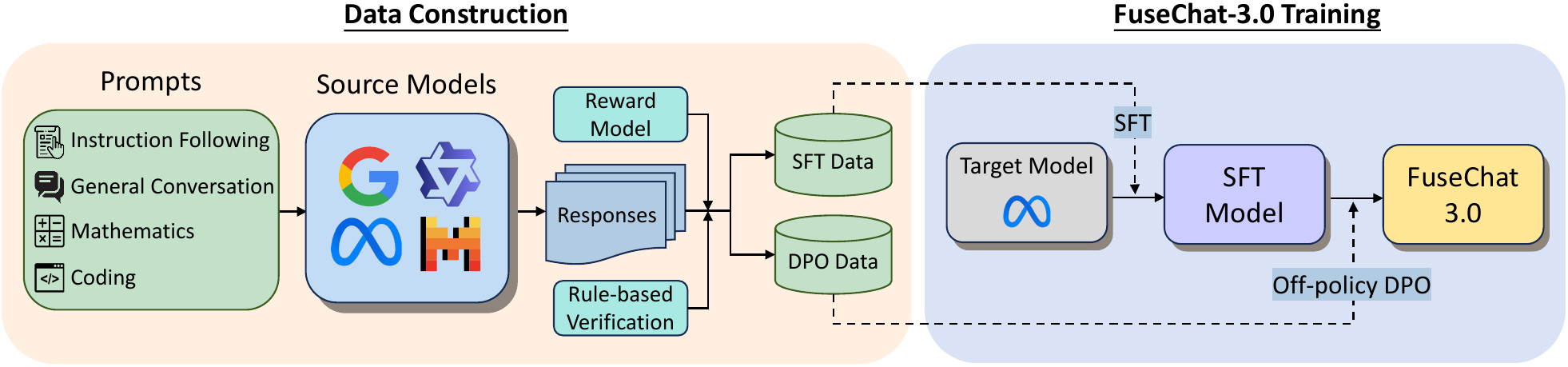}
	\caption{Overview of our proposed FuseChat-3.0 framework for implicit model fusion.}
	\label{fig: overview}
	\vspace{-0.3cm}
\end{figure*}

As illustrated in Figure~\ref{fig: overview}, our IMF framework follows a three-stage process. First, during the data construction stage, we generate multiple responses from various source models for each prompt. These responses are then evaluated using an external reward model for instruction-following tasks, while mathematics and coding responses are verified through rule-based methods. Next, we introduce a supervised fine-tuning (SFT) stage to tackle the challenges posed by distribution shifts when applying preference learning directly to heterogeneous LLM outputs~\citep{xu2024dpo, tajwar2024preference, zhou2024wpo}. Specifically, this stage fine-tunes the target models using the optimal response from the source models for each prompt. Finally, building on the SFT initialization, the Direct Preference Optimization (DPO) stage incorporates controlled preference signals from same-source response pairs to further fine-tune the target model. This structured approach enhances model robustness while reducing bias and variance caused by heterogeneous response data. 

To demonstrate the effectiveness of FuseChat-3.0, we conduct experiments across a diverse range of models and benchmarks. We utilize four prominent open-source LLMs, with parameter sizes ranging from 27B to 123B, as source models. For target models, we select five smaller, widely used models with sizes between 1B and 9B parameters. FuseChat-3.0 is evaluated on 14 well-established benchmarks spanning four core capability domains: instruction-following, general knowledge, mathematics, and coding. The results demonstrate that FuseChat-3.0 consistently outperforms its corresponding target LLMs, highlighting its effectiveness in facilitating the implicit fusion of heterogeneous LLMs across diverse tasks. Notably, when Llama-3.1-8B-Instruct is used as the target LLM, our fusion approach achieves an average improvement of 6.8 points across the 14 benchmarks. Moreover, we observe remarkable gains of 37.1 points on AlpacaEval-2 and 30.1 points on Arena-Hard, setting a new state-of-the-art performance for 8B-parameter LLMs. These results emphasize the robustness and efficacy of our approach in enhancing the capabilities of smaller target models.

\vspace{-0.1cm}
\section{Related Work}
\vspace{-0.1cm}
\label{related_works}
\paragraph{Collective LLMs}

With the prosperity of various open-source LLMs, it has become natural to explore ways to combine the strengths of these heterogeneous models to create more powerful systems. This section reviews existing approaches to combine heterogeneous LLMs, categorizing them into four main strategies: \emph{ensemble}, \emph{routing}, \emph{model merging}, and \emph{model fusion} methods.

\text{Ensemble methods} enhance model performance by aggregating the outputs of multiple LLMs. LLM-Blender~\citep{jiang2023llm} employs a pairwise ranker to select the top-$K$ outputs from various LLMs, which are then aggregated and refined by a sequence-to-sequence blender model. Mixture-of-Agents (MoA)~\citep{wang2024mixture} organizes LLM agents in a hierarchical structure, where each layer refines and combines the outputs from the preceding layer. Beyond sequence-level aggregation, \cite{xu2024bridging} introduced a token-level ensembling method that merges LLM distributions at each decoding step using a global alignment matrix. Although effective, ensemble methods are computationally expensive due to the simultaneous activation of multiple LLMs during inference.

\text{Routing methods} optimize efficiency by selectively invoking specific LLMs based on input characteristics, thereby reducing the computational cost of ensemble approaches. Hybrid-LLM~\citep{ding2024hybrid} trains a BERT-based router on synthetic preference labels generated using BARTScore~\citep{yuan2021bartscore} to dynamically assign queries to either a small or large model, depending on predicted query complexity and the desired output quality. RouteLLM~\citep{ong2025routellm} adopts a similar approach but utilizes real human preference labels from Chatbot Arena~\citep{chiang2024chatbot} to train the router. While these methods reduce costs and maintain high-quality responses with minimal latency, their reliance on task-specific router training limits their ability to generalize to new scenarios.

\text{Model merging methods}~\citep{wortsman2022model} aim to directly combine the parameters of multiple models into a single unified model. 
While enhancing robustness and generalization, they are limited to homogeneous model families.
\cite{xu2024training} introduced a strategy for merging models with differing depths and widths by aligning cohesive layer groups and applying elastic neuron zipping to project weights into a common space. While this approach allows for the merging of heterogeneous models, it encounters scalability challenges due to high computational complexity with large models.

\text{Model fusion methods} seek to consolidate the capabilities of multiple source models into a single target model and can be categorized as either explicit or implicit. Explicit model fusion (EMF) techniques, such as FuseLLM~\citep{wan2024knowledge} and FuseChat~\citep{wan2024fusechatknowledgefusionchat}, transfer knowledge from multiple source LLMs to a target LLM through multi-teacher knowledge distillation. FuseChat implements a two-stage process: first, it performs pairwise fusion between each source model and a pivot model; next, it merges the resulting homologous models in parameter space. While EMF methods adapt to various architectures and model sizes, they face challenges like vocabulary alignment and merging distribution matrices. To address these issues, WRPO~\citep{yang2025weightedreward} introduces implicit model fusion (IMF), which uses source model responses as auxiliary signals during on-policy preference optimization. WRPO progressively shifts optimization from the target model's on-policy outputs to high-reward off-policy responses from the source models.

\vspace{-0.2cm}
\paragraph{Preference Alignment}
Aligning large language models (LLMs) with human preferences is crucial for their effectiveness. Reinforcement learning from human feedback (RLHF)~\citep{christiano2017deep,ouyang2022training} is a widely used approach for this purpose. However, RLHF relies on complex reinforcement learning techniques like proximal policy optimization (PPO)~\citep{schulman2017proximal}, which can be challenging to implement and prone to instability during training. To address these challenges, Direct Preference Optimization (DPO)~\citep{Rafailov2023DirectPO} offers a simplified alternative by directly optimizing the policy model based on a closed-form implicit reward function derived from human preference data. This approach simplifies implementation, reduces computational costs, enhances training stability, and maintains strong alignment with human preferences.
\vspace{-0.1cm}
\section{FuseChat-3.0 Dataset}
\vspace{-0.1cm}
The data construction process is pivotal in enabling the implicit model fusion (IMF) technique utilized by FuseChat-3.0. This section outlines the methods used for prompt selection, response sampling, and data construction, highlighting the rationale behind our design choices.
\vspace{-0.1cm}
\subsection{Prompt Selection}
\vspace{-0.1cm}
Our primary objective is to curate a diverse dataset that enhances the comprehensive capabilities of target LLMs across multiple domains, including instruction following, mathematics, coding, and Chinese language proficiency. To accomplish this, we strategically select data from reputable open-source community datasets and implement targeted filtering and preprocessing techniques. Below is an overview of the key datasets and filtering criteria employed:
\begin{itemize}
    \item \textbf{Instruction Following}: The dataset is selected from multiple sources, including UltraFeedback~\citep{Cui2024UltraFeedbackBL}, Magpie-Pro-DPO~\citep{xu2025magpie}, and HelpSteer2~\citep{wang2024helpsteer2}. To maintain focus on general conversation, we systematically excluded code and mathematics-related data, ultimately assembling 80,907 distinct samples.
    \item \textbf{Mathematics}: We incorporate mathematical prompts from OpenMathInstruct-2~\citep{openmathinstruct2}, which yields approximately 52,000 unique entries with validated solutions.
    \item \textbf{Coding}: Our coding dataset is constructed from LeetCode\footnote{\url{https://huggingface.co/datasets/greengerong/leetcode}} problems and Self-Oss-Instruct-SC2~\citep{self-oss-instruct}. We specifically select problems that include comprehensive test cases for validation purposes, resulting in a final collection of 16,005 samples.
    \item \textbf{Chinese Language}: The Chinese language dataset is created by integrating Alpaca-GPT4-Zh~\citep{peng2023instruction} and Magpie-Qwen2-Pro-Zh~\citep{xu2025magpie}. After filtering out code and mathematics-related content, we obtain approximately 10,000 high-quality samples that specifically focus on Chinese dialogues.
\end{itemize}

\vspace{-0.1cm}
\subsection{Response Sampling}
\vspace{-0.1cm}
For each prompt within the curated datasets, we sample responses primarily from four prominent source LLMs: Gemma-2-27B-it~\citep{team2024gemma}, Mistral-Large-Instruct-2407~\citep{Jiang2023Mistral7}, Qwen-2.5-72B-Instruct~\citep{yang2024qwen25}, and Llama-3.1-70B-Instruct~\citep{dubey2024llama}. The response sampling strategy varies based on the specific domain:
\begin{itemize}
    \item \textbf{Instruction Following}: For each prompt, we sample five responses from each source model.
    \item \textbf{Mathematics}: We maintain the same sampling methodology as in the Instruction Following category. Additionally, we incorporate responses generated by Llama-3.1-405B-Instruct~\citep{dubey2024llama} and Qwen-2.5-Math-72B-Instruct~\citep{yang2024qwen25math} to enhance the correctness of our mathematical solution pool.
    \item \textbf{Coding}: To ensure a broader range of coding solutions, we collect eight distinct responses from each source model per prompt.
    \item \textbf{Chinese Language}: We utilize Qwen-2.5-72B-Instruct for response generation in this category given its specialized optimization for the Chinese language.
\end{itemize}
We use vLLM\footnote{\url{https://github.com/vllm-project/vllm}}~\citep{kwon2023efficient} as our inference back-end and sample responses multiple times with different random seeds. The sampling parameters for the different source LLMs are detailed as follows: For Gemma-2-27B-it, Mistral-Large-Instruct-2407, and Llama-3.1-70B-Instruct, we use a temperature of 0.8 and a top-p value of 0.95. For Qwen-2.5-(Math)-72B-Instruct, we set the temperature to 0.7, the top-p value to 0.8, and apply a repetition penalty of 1.05.

\vspace{-0.1cm}
\subsection{Preference Pairs}
\vspace{-0.1cm}
Unlike previous methods~\citep{tunstall2024zephyr,lambert2024t,yang2025weightedreward}, which construct preference pairs from different models with varying output styles, FuseChat-3.0 leverages the best and worst responses generated by the same source model to construct each preference pair and optimize the target model. This intra-model pairing eliminates the reward bias associated with heterogeneous response styles, prevents reward hacking, and offers a more controlled preference signal.
The data construction process varies based on the specific domain. For instruction-following and conversational data, we employ an external reward model to evaluate the sampled responses. In contrast, responses in mathematics and coding domains are verified through rule-based systems. We show the specific procedures for each domain as below:

\begin{itemize}
    \item \textbf{Instruction Following}: To assign reward model (RM) scores to the five responses generated by each source model, we employ ArmoRM-LLaMA3-8B-v0.1~\citep{ArmoRM} for annotation. We then divide the annotated data into SFT and DPO datasets using a 4:6 ratio. During the SFT phase, we select the highest RM-scoring response for each prompt across all source models for initial training. In the DPO phase, preference pairs are constructed by pairing responses from the same source model. Responses with the highest RM scores within each model are labeled as positive samples, while those with the lowest scores are treated as negative samples. We then use pairs where the positive responses have the highest RM scores to fine-tune the target model. To ensure the quality of preference signals, we impose a constraint on the RM score difference between positive and negative samples when constructing preference pairs, limiting it to the range of 0.01 to 0.1. This measure aims to maintain distinguishability between chosen and rejected samples, thereby minimizing potential interference to the optimization process caused by small differences or annotation noise, and thus improving the robustness and convergence of training.
    \item \textbf{Mathematics}: Responses from all source models are initially evaluated for correctness by comparing extracted answers against gold labels. This assessment is complemented by RM scores provided by ArmoRM-LLaMA3-8B-v0.1, ensuring both factual accuracy and adherence to stylistic preferences. We then strategically partition the dataset into SFT and DPO partitions. The SFT phase incorporates responses that are both correct and exhibit the highest RM scores. This selection strategy ensures that the fine-tuning process is grounded in high-quality responses closely aligned with the desired task outcomes. For the DPO phase, we construct paired samples from the same source model, as described above. Positive samples consist of correct answers with the highest RM scores, while negative samples comprise incorrect answers with the lowest RM scores. 
    \item \textbf{Coding}: We employ a verification method comprising correctness scores and RM scores for coding evaluation. Correctness scores assess whether the generated code passes both static analysis checks and provided test cases, ensuring functional accuracy. The RM scores are used for preference evaluation, gauging the overall quality of responses based on predefined criteria. During the SFT phase, we select responses that not only pass all test cases but also achieve the highest RM scores. This rigorous selection ensures that the model is fine-tuned on exemplary code meeting both correctness and preference standards. In the DPO phase, we compare positive samples—high-scoring responses that pass all evaluations—with negative samples—low-scoring responses that fail the tests, both originating from the same source model. This comparison aims to optimize the model's ability to prefer higher-quality and functionally correct code during training. Critically, we exclude instances where all model responses fail to meet the testing criteria. This exclusion is implemented to maintain the integrity of the evaluation process, as such cases do not provide meaningful data for assessing and improving the model's performance.
    \item \textbf{Chinese Language}: We exclusively utilize responses sampled from Qwen-2.5-72B-Instruct during the SFT phase, due to its strong performance in the Chinese language. This focused approach aims to maximize the transfer of high-quality Chinese language capabilities to the target model. Since no suitable reward models were available for this dataset, we opted to omit the DPO phase.
\end{itemize}

Our final dataset \(\mathcal{D}\) consists of 158,667 entries, with 94,539 allocated to the SFT phase (\(\mathcal{D}_\text{SFT}\)) and 64,128 preference pairs for the DPO phase (\(\mathcal{D}_\text{DPO}\)). A summary of the dataset composition is provided in Table~\ref{tab:dataset_composition}. Refer to Appendix~\ref{appendix:model_dataset_details} for further details on the open-source models and datasets. 

\begin{table}[t]
\caption{The constitution of FuseChat-3.0 dataset in SFT phase and DPO phase. As no suitable reward models were available for Chinese, we used all samples for SFT and omitted the DPO phase.}
\centering
\setlength\tabcolsep{5pt}
{\small
\begin{tabular}{ll l L{1.7cm} L{1.7cm}}
\toprule
\textbf{Category} & \textbf{Dataset}  & \textbf{Count}  & \textbf{\#\(\mathcal{D}_\text{SFT}\)}  & \textbf{\#\(\mathcal{D}_\text{DPO}\)}  \\
\midrule
Instruction Following & UltraFeedback & 51,098 & 20,439 & 30,659 \\
& Magpie-Pro-DPO & 20,374 & 8,149 & 12,225 \\
& HelpSteer2 & 9,435 & 3,774 & 5,661 \\ \midrule
Mathematics & OpenMathInstruct-2 & 51,803 & 40,188 & 11,615 \\ \midrule
Coding & LeetCode & 3,113 & 1,877 & 1,236 \\
& Self-Oss-Instruct-SC2 & 12,892 & 10,160 & 2,732 \\ \midrule
Chinese Language & Alpaca-GPT4-Zh & 2,471 & 2,471 & 0 \\
& Magpie-Qwen2-Pro-Zh & 7,481 & 7,481 & 0 \\ \midrule
\textit{Total} &  & 158,667 & 94,539 & 64,128 \\
\bottomrule
\end{tabular}}
\vspace{-0.2cm}
\label{tab:dataset_composition}
\end{table}

\vspace{-0.15cm}
\section{Training Recipe}
\vspace{-0.1cm}
This section details the two-stage training pipeline designed for FuseChat-3.0. The first stage involves supervised fine-tuning (SFT) to address distributional discrepancies between the target and source LLMs. 
Building on the SFT initialization, 
the second stage employs preference learning, specifically Direct Preference Optimization (DPO), to learn preferences from multiple source LLMs.

\vspace{-0.15cm}
\subsection{Supervised Fine-Tuning}
\vspace{-0.1cm}
Given a prompt \(x_i\) and its corresponding output \(y_i\) of length \(N\) from the fine-tuning dataset \(\mathcal{D}_\text{SFT}\), we denote the sequence preceding the \(t\)-th token in the output as \(y_{i, <t} = (y_{i, 1}, y_{i, 2}, \dots, y_{i, t-1})\). The SFT objective for a language model with parameters \(\theta\) is to minimize the negative log-likelihood:
\begin{equation}
\label{eqn:token_level_clm}
\mathcal{L}_{\text{SFT}}(\theta) = -\mathbb{E}_{(x_i, y_i) \sim \mathcal{D}_\text{SFT}}\left[\sum_{t=1}^{N} \log p_\theta(y_{i, t} \mid y_{i, <t}, x_i)\right],
\end{equation}
where \(p_\theta(y_{i, t} \mid y_{i, <t}, x_i)\) represents the model's predicted probability for the \(t\)-th token \(y_{i,t}\) in \(y_i\), conditioned on the prompt \(x_i\) and the preceding tokens \(y_{i, <t}\).

In our experiments, we use the Llama-Factory library\footnote{\url{https://github.com/hiyouga/LLaMA-Factory}}~\citep{llamafactory} to implement the fine-tuning. For all target models, we perform fine-tuning for 3 epochs, with a batch size of 128 and a maximum sequence length of 2048 tokens. A cosine learning rate schedule with a warmup ratio of 0.1 is employed. The learning rates for different models are shown in Table~\ref{tab:trainig_hyperparameters}.

\vspace{-0.15cm}
\subsection{Direct Preference Optimization}
\vspace{-0.1cm}
 The DPO objective directly optimizes a policy to align with human preferences by leveraging a supervised learning objective on human-labeled preference data. DPO reformulates the reward function to yield a closed-form solution for the optimal policy $\pi^*$. The reparameterized optimal reward function $r^*(x, y)$ is denoted as:
\begin{equation}
\label{eq:dpo_reward_fuction}
r^*(x, y) = \beta \log \frac{\pi^*(y \mid x)}{\pi_{\text{ref}}(y \mid x)} + \beta \log Z(x),
\end{equation}
where $Z(x)$ is the partition function, $\pi_{\text{ref}}$ denotes the reference policy, and $\beta$ is a hyperparameter controlling the deviation from $\pi_{\text{ref}}$. Given a preference dataset \(\mathcal{D}_\text{DPO}\) consisting of triplets \((x, y_w, y_l)\), where $y_w$ and $y_l$ are the preferred and dispreferred completions for a given prompt $x$, the preference probability for $y_w$ over $y_l$ is modeled using the Bradley-Terry model~\citep{Bradley1952RankAO}:
\begin{equation}
\label{eq:objective}
p^*(y_w \succ y_l \mid x) = \sigma\left(\beta \log \frac{\pi^*(y_w \mid x)}{\pi_{\text{ref}}(y_w \mid x)} - \beta \log \frac{\pi^*(y_l \mid x)}{\pi_{\text{ref}}(y_l \mid x)}\right).
\end{equation}
For a parameterized policy $\pi_\theta$, the corresponding maximum likelihood objective is:
\begin{equation}
\label{eq:dpo}
\mathcal{L}_{\text{DPO}}(\pi_\theta; \pi_{\text{ref}}) = -\mathbb{E}_{(x, y_w, y_l) \sim \mathcal{D}_\text{DPO}}\left[\log \sigma\left(\beta \log \frac{\pi_\theta(y_w \mid x)}{\pi_{\text{ref}}(y_w \mid x)} - \beta \log \frac{\pi_\theta(y_l \mid x)}{\pi_{\text{ref}}(y_l \mid x)}\right)\right].
\end{equation}
In this work, we explore the application of length normalization during DPO training, a variant proposed in SimPO~\citep{meng2024simpo} and proven to be effective in T\"ulu 3 \citep{lambert2024t}. Specifically, the length-normalized DPO objective is defined as:
\begin{equation}
\label{eq:lndpo}
\mathcal{L}_{\text{LN-DPO}}(\pi_\theta; \pi_{\text{ref}}) = -\mathbb{E}_{(x, y_w, y_l) \sim \mathcal{D}_\text{DPO}}\left[\log \sigma\left(\frac{\beta}{|y_w|} \log \frac{\pi_\theta(y_w \mid x)}{\pi_{\text{ref}}(y_w \mid x)} - \frac{\beta}{|y_l|} \log \frac{\pi_\theta(y_l \mid x)}{\pi_{\text{ref}}(y_l \mid x)}\right)\right],
\end{equation}
where $|y_w|$ and $|y_l|$ denote the lengths of the preferred and dispreferred completions, respectively.

In our experiments, we utilize the alignment-handbook\footnote{\url{https://github.com/huggingface/alignment-handbook}} as the training framework for DPO. All post-SFT target models undergo training for one epoch with a batch size of 128 and a maximum sequence length of 2048. A cosine learning rate schedule with a warmup ratio of 0.1 is used. Checkpoints are saved every 100 steps, and the best checkpoint from the last two is selected. Hyperparameter configurations for different models are detailed in Table~\ref{tab:trainig_hyperparameters}.

\begin{table}[!t]
\centering
\vspace{-0.1cm}
\caption{Learning rates and hyperparameters for different target models during the SFT and DPO stages.}
\label{tab:trainig_hyperparameters}
\resizebox{0.99\linewidth}{!}{
    \begin{tabular}{lcccc}
    \toprule
    \textbf{Target Model} & \textbf{SFT Learning Rate} & \textbf{DPO Learning Rate} & \textbf{DPO \(\beta\)} & \textbf{DPO Loss Type} \\
    \midrule
    Llama-3.1-8B-Instruct & $5 \times 10^{-6}$ & $8 \times 10^{-7}$ & 10 & $\mathcal{L}_{\text{LN-DPO}}$ \\
    Qwen-2.5-7B-Instruct & $2 \times 10^{-6}$ & $3 \times 10^{-7}$ & 0.01 & $\mathcal{L}_{\text{DPO}}$ \\
    Gemma-2-9B-it & $2 \times 10^{-6}$ & $5 \times 10^{-7}$ & 0.01 & $\mathcal{L}_{\text{DPO}}$ \\
    Llama-3.2-(3/1)B-Instruct & $5 \times 10^{-6}$ & $1 \times 10^{-6}$ & 10 & $\mathcal{L}_{\text{LN-DPO}}$ \\
    \bottomrule
    \end{tabular}
}
\vspace{-0.2cm}
\end{table}

\vspace{-0.15cm}
\section{Evaluation}
\vspace{-0.1cm}
\subsection{Evaluation Benchmarks}
\vspace{-0.1cm}
To showcase the superior performance of FuseChat-3.0, we conduct a comprehensive evaluation across multiple domains, including instruction following, question answering, general reasoning, mathematics, and coding. Specifically, we utilize 14 well-established benchmarks, which are categorized into four distinct groups:
\begin{itemize}
    \item \textbf{Instruction-Following Tasks}: AlpacaEval-2~\citep{AlpacaEval}, Arena-Hard~\citep{arenahard2024}, MT-Bench~\citep{zheng2023judging}, AlignBench v1.1~\citep{liu2023alignbench}.
    \item \textbf{General Tasks}: LiveBench-0831~\citep{white2025livebench}, MMLU-Pro~\citep{wang2024mmlupro}, MMLU-redux~\citep{gema2024we}, GPQA-Diamond~\citep{rein2023gpqa}.
    \item \textbf{Mathematics Tasks}: GSM8K~\citep{cobbe2021gsm8k}, MATH~\citep{hendrycks2021math}, AMC 23~\citep{yang2024qwen25math}.
    \item \textbf{Coding Tasks}: HumanEval~\citep{chen2021evaluating}, MBPP~\citep{austin2021program}, LiveCodeBench 2408-2411~\citep{jain2024livecodebench}.
\end{itemize}
Further details about these benchmarks are provided in Appendix~\ref{appendix:benchhmark_details}.

\begin{table}[t]
\caption{Overall results of FuseChat-3.0, with Llama-3.1-8B-Instruct, Llama-3.2-3B-Instruct, and Llama-3.2-1B-Instruct as target models. \textbf{\small Bold} denotes the best performance on each benchmark.}
\vspace{-0.1cm}
\centering
\setlength\tabcolsep{5pt}
\adjustbox{max width=\linewidth}{
\begin{NiceTabular}{@{}ll|C{20pt}C{20pt}C{42pt}|C{20pt}C{20pt}C{42pt}|C{20pt}C{20pt}C{42pt}@{}}
\toprule
\multirow{2}{*}{\textbf{Category}} & \multirow{2}{*}{\textbf{Benchmark}} & \multicolumn{3}{c}{\textbf{Llama-3.1-8B-Instruct}} & \multicolumn{3}{c}{\textbf{Llama-3.2-3B-Instruct}} & \multicolumn{3}{c}{\textbf{Llama-3.2-1B-Instruct}} \\ \cline{3-11}
 & & \rule{0pt}{11pt}\textbf{Base} & \rule{0pt}{11pt}\textbf{SFT} & \rule{0pt}{11pt}\cellcolor{mylightblue}\textbf{FuseChat} & \rule{0pt}{11pt}\textbf{Base} & \rule{0pt}{11pt}\textbf{SFT} & \rule{0pt}{11pt}\cellcolor{mylightblue}\textbf{FuseChat} & \rule{0pt}{11pt}\textbf{Base} & \rule{0pt}{11pt}\textbf{SFT} & \rule{0pt}{11pt}\cellcolor{mylightblue}\textbf{FuseChat} \\ \midrule
    & AlpacaEval-2~$_\text{(LC \%)}$ & 28.3 & 41.3 & \cellcolor{mylightblue}\textbf{65.4} & 21.4 & 31.1 & \cellcolor{mylightblue}\textbf{54.0} & 9.7 & 14.0 & \cellcolor{mylightblue}\textbf{25.3} \\
    Instruction & Arena-Hard~$_\text{(WR \%)}$ & 28.1 & 38.7 & \cellcolor{mylightblue}\textbf{58.2} & 16.6 & 21.3 & \cellcolor{mylightblue}\textbf{30.2} & 5.1 & 6.0 & \cellcolor{mylightblue}\textbf{8.6} \\
    Following & MT-Bench & 8.4 & 8.5 & \cellcolor{mylightblue}\textbf{9.0} & 6.9 & 7.3 & \cellcolor{mylightblue}\textbf{7.7} & 4.7 & 5.2 & \cellcolor{mylightblue}\textbf{5.7} \\
    & AlignBench$_\text{v1.1}$  & 4.6 & 6.3 & \cellcolor{mylightblue}\textbf{6.7} & 3.8 & 5.5 & \cellcolor{mylightblue}\textbf{5.9} & 2.9 & 3.9 & \cellcolor{mylightblue}\textbf{4.3} \\ \midrule
    \multirow{4}{*}{General} & LiveBench$_\text{0831}$  & 27.6 & 30.2 & \cellcolor{mylightblue}\textbf{32.0} & 23.4 & 24.5 & \cellcolor{mylightblue}\textbf{24.9} & 14.0 & 13.9 & \cellcolor{mylightblue}\textbf{15.8} \\
    & MMLU-Pro~$_\text{(0 shot, CoT)}$ & \textbf{50.0} & 47.8 & \cellcolor{mylightblue}49.2 & 39.3 & \textbf{40.3} & \cellcolor{mylightblue}40.3 & \textbf{22.3} & 21.5 & \cellcolor{mylightblue}21.3 \\
    & MMLU-redux~$_\text{(0 shot, CoT)}$ & 67.2 & 68.4 & \cellcolor{mylightblue}\textbf{69.2} & 58.5 & 58.2 & \cellcolor{mylightblue}\textbf{59.0} & \textbf{43.7} & 40.3 & \cellcolor{mylightblue}41.6 \\
    & GPQA-Diamond~$_\text{(0 shot, CoT)}$ & 33.8 & \textbf{37.9} & \cellcolor{mylightblue}34.9 & 29.8 & 33.3 & \cellcolor{mylightblue}\textbf{33.8} & 21.2 & \textbf{25.3} & \cellcolor{mylightblue}24.2 \\ \midrule
    \multirow{3}{*}{Mathematics} & GSM8K~$_\text{(0 shot, CoT)}$ & 85.9 & 87.0 & \cellcolor{mylightblue}\textbf{88.0} & 82.0 & \textbf{82.8} & \cellcolor{mylightblue}82.0 & 46.3 & \textbf{55.6} & \cellcolor{mylightblue}54.5 \\
    & MATH~$_\text{(0 shot, CoT)}$ & 50.7 & 54.7 & \cellcolor{mylightblue}\textbf{55.2} & 51.4 & 52.9 & \cellcolor{mylightblue}\textbf{53.1} & 32.7 & \textbf{34.7} & \cellcolor{mylightblue}33.6 \\
    & AMC 23~$_\text{(0 shot, CoT)}$ & 25.0 & 30.0 & \cellcolor{mylightblue}\textbf{37.5} & 22.5 & 20.0 & \cellcolor{mylightblue}\textbf{35.0} & 17.5 & 15.0 & \cellcolor{mylightblue}\textbf{20.0} \\ \midrule
    \multirow{3}{*}{Coding} & HumanEval~$_\text{(0 shot)}$ & 69.5 & 69.5 & \cellcolor{mylightblue}\textbf{71.3} & 61.0 & \textbf{62.8} & \cellcolor{mylightblue}60.4 & 39.6 & 36.6 & \cellcolor{mylightblue}\textbf{40.2} \\ 
    & MBPP~$_\text{(0 shot)}$ & \textbf{75.4} & 71.4 & \cellcolor{mylightblue}72.0 & \textbf{68.5} & 67.5 & \cellcolor{mylightblue}67.5 & \textbf{49.5} & 42.1 & \cellcolor{mylightblue}46.6 \\
    & LiveCodeBench$_\text{2408-2411}$ & 12.3 & 12.6 & \cellcolor{mylightblue}\textbf{13.1} & 8.3 & 7.1 & \cellcolor{mylightblue}\textbf{9.0} & - & - & \cellcolor{mylightblue}- \\
    \midrule
    \multicolumn{2}{c}{Average} & 40.5 & 43.2 & \cellcolor{mylightblue}\textbf{47.3} & 35.2 & 36.8 & \cellcolor{mylightblue}\textbf{40.2} & 23.8 & 24.2 & \cellcolor{mylightblue}\textbf{26.3} \\
\bottomrule
\end{NiceTabular}}
\vspace{-0.2cm}
\label{tab:llama_all_performance}
\end{table}

\subsection{Overall Results}
In Table \ref{tab:llama_all_performance} and Table \ref{tab:gemma_qwen_performance}, we present the overall results of our FuseChat-3.0, with Llama-3.1-8B-Instruct, Llama-3.2-3B-Instruct, Llama-3.2-1B-Instruct, Qwen-2.5-7B-Instruct, and Gemma-2-9B-it as the target models. Based on the experimental results, we identify several key insights.

\begin{table}[t]
\caption{Overall results of our FuseChat-3.0 framework, with Qwen-2.5-7B-Instruct and Gemma-2-9B-it served as target models. \textbf{\small Bold} indicates the best performance on each benchmark and of each target model.}
\vspace{-0.1cm}
\centering
\setlength\tabcolsep{5pt}
\adjustbox{max width=0.80\linewidth}{
\begin{NiceTabular}{@{}ll|C{22pt}C{22pt}C{42pt}|C{22pt}C{22pt}C{42pt}@{}}
\toprule
\multirow{2}{*}{\textbf{Category}} & \multirow{2}{*}{\textbf{Benchmark}} & \multicolumn{3}{c}{\textbf{Qwen-2.5-7B-Instruct}} & \multicolumn{3}{c}{\textbf{Gemma-2-9B-it}} \\ \cline{3-8}
& & \rule{0pt}{11pt}\textbf{Base} & \rule{0pt}{11pt}\textbf{SFT} & \rule{0pt}{11pt}\cellcolor{mylightblue}\textbf{FuseChat} & \rule{0pt}{11pt}\textbf{Base} & \rule{0pt}{11pt}\textbf{SFT} & \rule{0pt}{11pt}\cellcolor{mylightblue}\textbf{FuseChat} \\ \midrule
    & AlpacaEval-2~$_\text{(LC \%)}$ & 33.2 & 34.2 & \cellcolor{mylightblue}\textbf{63.6} & 51.1 & 49.8 & \cellcolor{mylightblue}\textbf{70.2} \\
    Instruction & Arena-Hard~$_\text{(WR \%)}$ & 50.7 & 45.2 & \cellcolor{mylightblue}\textbf{61.4} & 40.8 & 44.5 & \cellcolor{mylightblue}\textbf{63.4} \\
    Following & MT-Bench & 8.4 & 8.5 & \cellcolor{mylightblue}\textbf{9.0} & 8.5 & \textbf{8.7} & \cellcolor{mylightblue}8.6 \\
    & AlignBench$_\text{v1.1}$  & 7.5 & 7.4 & \cellcolor{mylightblue}\textbf{7.6} & 7.0 & 7.1 & \cellcolor{mylightblue}\textbf{7.4} \\ \midrule
    \multirow{4}{*}{General} & LiveBench$_\text{0831}$  & \textbf{35.4} & 33.7 & \cellcolor{mylightblue}33.2 & 31.6 & \textbf{33.3} & \cellcolor{mylightblue}33.2 \\
    & MMLU-Pro~$_\text{(0 shot, CoT)}$ & \textbf{54.1} & 51.7 & \cellcolor{mylightblue}53.0 & 50.5 & 52.5 & \cellcolor{mylightblue}\textbf{52.9} \\
    & MMLU-redux~$_\text{(0 shot, CoT)}$ & \textbf{75.1} & 72.7 & \cellcolor{mylightblue}74.4 & 72.8 & 72.8 & \cellcolor{mylightblue}\textbf{73.7} \\
    & GPQA-Diamond~$_\text{(0 shot, CoT)}$ & 34.9 & \textbf{38.4} & \cellcolor{mylightblue}33.8 & \textbf{39.4} & 33.3 & \cellcolor{mylightblue}35.4 \\ \midrule
    \multirow{3}{*}{Mathematics} & GSM8K~$_\text{(0 shot, CoT)}$ & 91.7 & \textbf{92.3} & \cellcolor{mylightblue}91.7 & 88.5 & 90.5 & \cellcolor{mylightblue}\textbf{91.0} \\
    & MATH~$_\text{(0 shot, CoT)}$ & \textbf{75.0} & 72.7 & \cellcolor{mylightblue}73.6 & 49.6 & \textbf{58.0} & \cellcolor{mylightblue}57.8 \\
    & AMC 23~$_\text{(0 shot, CoT)}$ & 52.5 & 45.0 & \cellcolor{mylightblue}\textbf{57.5} & 20.0 & 27.5 & \cellcolor{mylightblue}\textbf{35.0} \\ \midrule
    \multirow{3}{*}{Coding} & HumanEval~$_\text{(0 shot)}$ & \textbf{85.4} & 81.7 & \cellcolor{mylightblue}79.9 & \textbf{67.1} & 65.9 & \cellcolor{mylightblue}64.0 \\ 
    & MBPP~$_\text{(0 shot)}$ & 80.2 & \textbf{84.1} & \cellcolor{mylightblue}83.1 & \textbf{75.1} & 70.6 & \cellcolor{mylightblue}71.7 \\
    & LiveCodeBench$_\text{2408-2411}$ & 15.8 & 17.3 & \cellcolor{mylightblue}\textbf{18.9} & \textbf{11.9} & 11.0 & \cellcolor{mylightblue}10.1 \\
    \midrule
    \multicolumn{2}{c}{Average} & 50.0 & 48.9 & \cellcolor{mylightblue}\textbf{52.9} & 43.9 & 44.7 & \cellcolor{mylightblue}\textbf{48.2} \\
\bottomrule
\end{NiceTabular}}
\vspace{-0.3cm}
\label{tab:gemma_qwen_performance}
\end{table}

Firstly, when using Llama-3.1-8B-Instruct as the target model, our FuseChat-3.0 achieves an average performance improvement of 6.8 points across 14 benchmarks compared to the original Llama-3.1-8B-Instruct model. Notably, it shows significant gains of 37.1 and 30.1 points on the instruction-following test sets AlpacaEval-2 and Arena-Hard, respectively, establishing new state-of-the-art performance for 8B LLMs. 
Moreover, FuseChat-3.0 improves performance in general knowledge tasks by an average of 1.6 points, and in mathematics by an average of 6.3 points, demonstrating comprehensive capability enhancement. However, we observe a slight decline of 0.3 average points in coding tasks, which may be attributed to the relatively limited coding data in our training set.

Secondly, the consistent improvements across different model sizes, ranging from 1B to 8B, for the Llama-3 series demonstrate the scalability of FuseChat-3.0. Notably, when applied to Llama-3.2-3B-Instruct, FuseChat-3.0 achieves an average score of 40.2, nearly matching Llama-3.1-8B-Instruct, which is 2.7 times larger. This highlights the potential of FuseChat-3.0 to enable smaller models to perform comparably to their larger counterparts, offering significant efficiency gains.

Finally, when applied to the stronger target model, Qwen-2.5-7B-Instruct, FuseChat-3.0 achieves an average improvement of 2.9 points, despite the higher baseline performance. While the relative improvement is smaller compared to the Llama-3 models, FuseChat-3.0 still boosts performance across several benchmarks. For instance, on AlpacaEval-2, FuseChat-3.0 boosts the score by 30.4 points. Similarly, on Arena-Hard, it achieves an improvement of 10.7 points. When using Gemma-2-9B-it as the target model, FuseChat-3.0 also delivers an average performance gain of 4.3 points. These results validate FuseChat-3.0 as an effective and scalable framework for enhancing capabilities through implicit knowledge fusion, achieving significant performance gains across model architectures and scales without requiring architectural modifications or massive computational resources.

\subsection{Comparison with T\"ulu 3}
\vspace{-0.2cm}
\begin{wrapfigure}{r}{0.4\textwidth}
    \vspace{-0.4cm}
    \centering
    \includegraphics[width=0.98\linewidth]{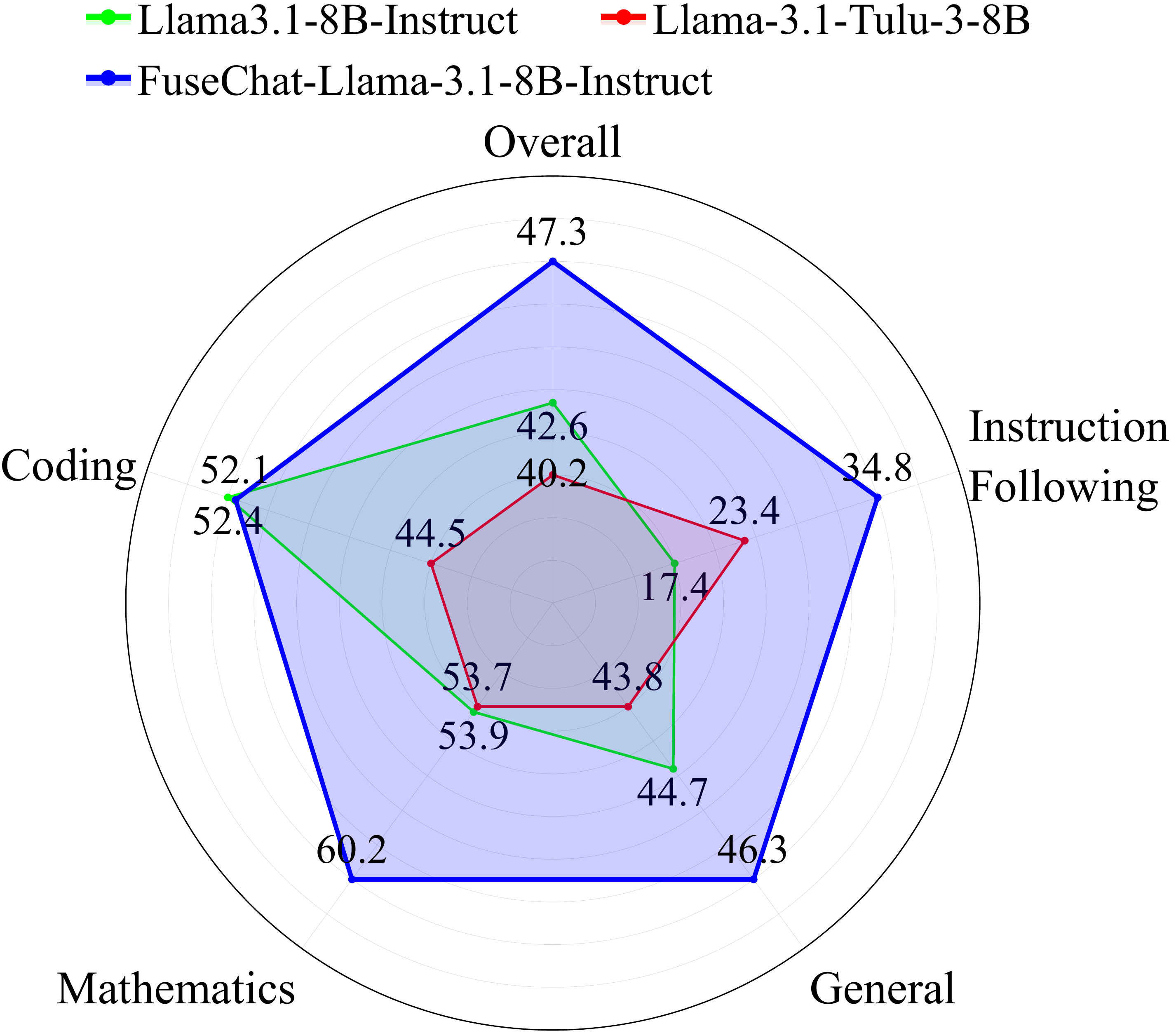}
    \caption{Comparison between FuseChat-Llama-3.1-8B-Instruct and Llama-3.1-Tulu-8B across different domains. Domain scores are obtained by averaging the scores of the benchmarks within that domain.}
    \label{fig:llama_radar}
    \vspace{-0.2cm}
\end{wrapfigure}

We conduct a comparative analysis between our FuseChat-3.0 framework and AllenAI's recently introduced T\"ulu 3~\citep{lambert2024t}. T\"ulu 3 employs an extensive training pipeline, which includes leveraging 22 models for data synthesis, GPT-4o for preference annotation, a training dataset exceeding 1.2 million data points, and a three-stage training pipeline including SFT, DPO, and PPO. 
As shown in Figure~\ref{fig:llama_radar}, our FuseChat-Llama-3.1-8B-Instruct model outperforms Llama-3.1-Tulu-8B across all evaluation categories. 
FuseChat-3.0 achieves an overall score of 47.3, exceeding T\"ulu 3 by 7.1 points. The performance gap is most pronounced in the instruction-following category, where FuseChat leads by 11.4 points. Superior performance is also demonstrated in the general, mathematics, and coding categories. 
These results suggest that while T\"ulu 3 benefits from a sophisticated training pipeline and a large dataset, our FuseChat-3.0 framework achieves better performance through efficient implicit model fusion. 
By rigorously curating each preference pair from the best and worst responses generated by the same source model, FuseChat-3.0 effectively mitigates reward bias and variance inherent in T\"ulu 3's approach, which constructs each preference pair using different models. These more controlled preference signals enable FuseChat-3.0 to achieve high performance with significantly lower computational cost during the preference optimization process.
\vspace{-0.2cm}
\subsection{Effectiveness of length-normalized DPO}
\vspace{-0.2cm}
\begin{wrapfigure}{r}{0.42\textwidth}
    \vspace{-0.4cm}
    \centering
    \includegraphics[width=0.999\linewidth]{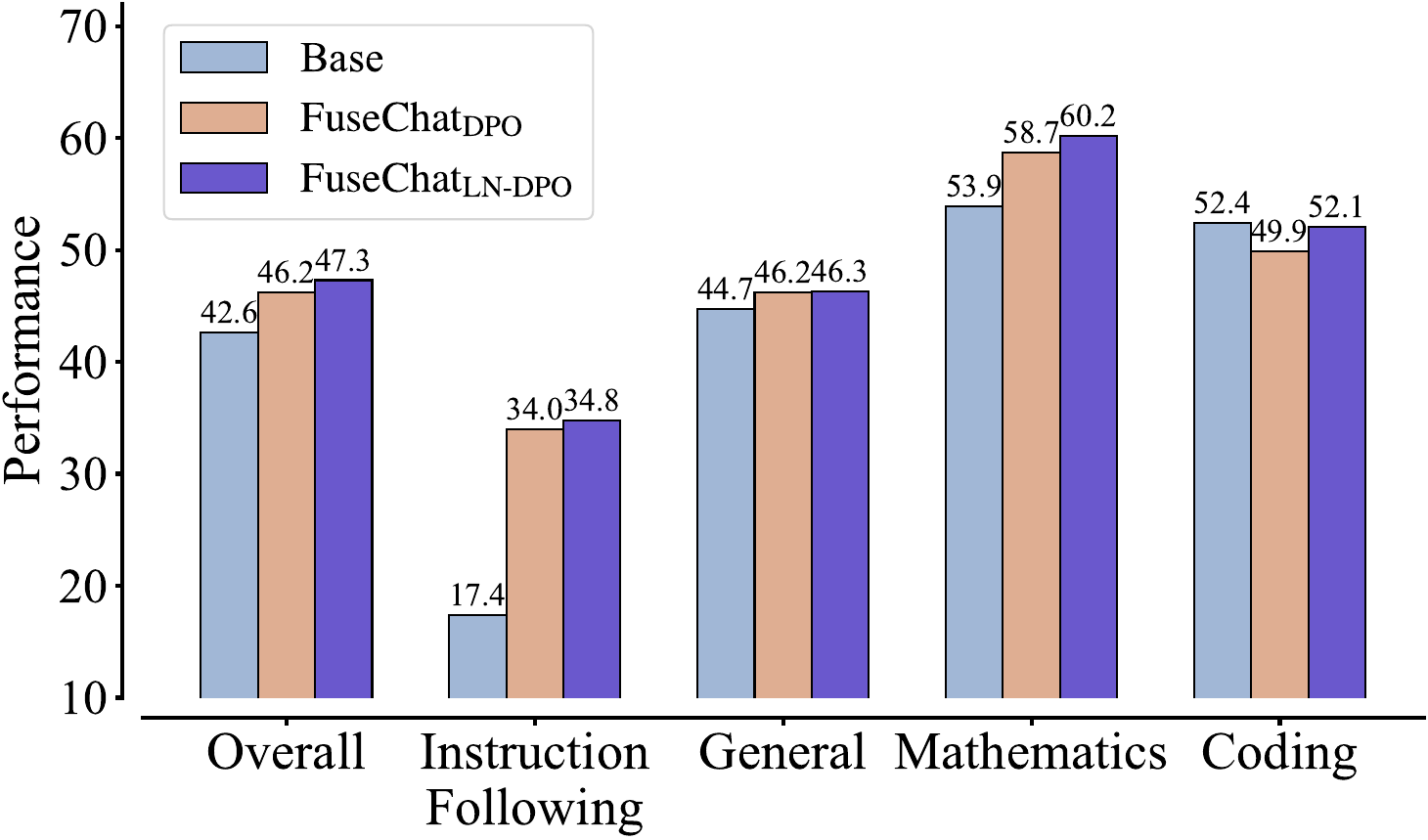}
        \vspace{-0.4cm}
    \caption{Comparison between length-normalized DPO and vanilla DPO in FuseChat-3.0. ``Base'' denotes Llama-3.1-8B-Instruct.}
    \label{fig:lndpo_bar}
    \vspace{-0.2cm}
\end{wrapfigure}

We investigate the impact of length normalization on the DPO stage of our FuseChat-3.0 and present the results in Figure~\ref{fig:lndpo_bar}. When using Llama-3.1-8B-Instruct as the target model, we show that length-normalized DPO (LN-DPO) generally leads to performance gains compared to both the standard DPO approach and the base model. Specifically, FuseChat$_\text{LN-DPO}$ achieves the highest overall score of 47.3, surpassing FuseChat$_\text{DPO}$'s 46.2 and the base model's 42.6. The most significant improvements due to length normalization are observed in the mathematics and coding categories. 
We attribute these substantial gains to the fact that length normalization effectively mitigates the length bias inherent in preference pairs. By reducing the model's tendency to favor longer responses, this technique ensures that the optimization process prioritizes correctness and completeness over unnecessary verbosity, which is particularly crucial in mathematics and coding domains requiring precise solutions. 
While LN-DPO also yields improvements in other categories, these gains are less substantial. These findings further reinforce that length normalization is a valuable technique for enhancing the performance of DPO-trained models, including FuseChat-3.0.
\vspace{-0.2cm}
\section{Conclusion}
\vspace{-0.2cm}
In this paper, we introduce FuseChat-3.0, a suite of large language models (LLMs) designed to integrate the strengths of heterogeneous source LLMs into more compact and efficient target models. To harness the diverse capabilities of these source models, we develop a specialized data construction protocol tailored to various tasks and domains. The FuseChat-3.0 training pipeline consists of a supervised fine-tuning (SFT) to align the target model with source model distributions, and a Direct Preference Optimization (DPO) stage to incorporate preferences from multiple source LLMs for further refinement. By leveraging four powerful open-source LLMs as source models and fine-tuning five smaller models as targets, FuseChat-3.0 consistently achieves substantial performance gains across 14 established benchmarks. These results highlight its effectiveness in transferring and integrating the diverse strengths of heterogeneous LLMs into smaller, more efficient models.

\bibliography{iclr2025_conference}
\bibliographystyle{iclr2025_conference}

\appendix
\newpage

\section{Details of Open-source Models and the Dataset}
\label{appendix:model_dataset_details}
In Table~\ref{tab:huggingface_models}, we provide the Huggingface repository names and links of the target LLMs, source LLMs, reward model, and the source of training datasets used in our experiments.

\begin{table}[ht]
\centering
\caption{Details of open-source models and datasets used in our experiments.}
\centering
    \resizebox{0.75\textwidth}{!}{
        \begin{tabular}{lc}
        \toprule
        \textbf{Name} & \textbf{Huggingface ID} \\ \midrule
        \multicolumn{2}{c}{\textbf{Target LLMs}} \\ \midrule
        Llama-3.1-8B-Instruct & \href{https://huggingface.co/meta-llama/Llama-3.1-8B-Instruct}{meta-llama/Llama-3.1-8B-Instruct} \\
        Gemma-2-9B-IT & \href{https://huggingface.co/google/gemma-2-9b-it}{google/gemma-2-9b-it}  \\
        Qwen-2.5-7B-Instruct & \href{https://huggingface.co/Qwen/Qwen2.5-7B-Instruct}{Qwen/Qwen2.5-7B-Instruct} \\
        Llama-3.2-3B-Instruct & \href{https://huggingface.co/meta-llama/Llama-3.2-3B-Instruct}{meta-llama/Llama-3.2-3B-Instruct} \\
        Llama-3.2-1B-Instruct & \href{https://huggingface.co/meta-llama/Llama-3.2-1B-Instruct}{meta-llama/Llama-3.2-1B-Instruct} \\ \midrule
        \multicolumn{2}{c}{\textbf{Source LLMs}} \\ \midrule
        Mistral-Large-Instruct-2407 & \href{https://huggingface.co/mistralai/Mistral-Large-Instruct-2407}{Mistral-Large-Instruct-2407} \\
        Gemma-2-27B-it & \href{https://huggingface.co/google/gemma-2-27b-it}{google/gemma-2-27b-it} \\
        Qwen-2.5-72B-Instruct & \href{https://huggingface.co/Qwen/Qwen2.5-72B-Instruct}{Qwen/Qwen2.5-72B-Instruct} \\
        Llama-3.1-70B-Instruct & \href{https://huggingface.co/meta-llama/Llama-3.1-70B-Instruct}{meta-llama/Llama-3.1-70B-Instruct} \\ \midrule
        \multicolumn{2}{c}{\textbf{Reward Model}} \\ \midrule
        ArmoRM-LLaMA3-8B-v0.1 & \href{https://huggingface.co/RLHFlow/ArmoRM-Llama3-8B-v0.1}{RLHFlow/ArmoRM-Llama3-8B-v0.1} \\ \midrule
        \multicolumn{2}{c}{\textbf{Datasets}} \\ \midrule
        UltraFeedback & \href{https://huggingface.co/datasets/princeton-nlp/llama3-ultrafeedback-armorm}{princeton-nlp/llama3-ultrafeedback-armorm}  \\
        Magpie-Pro-DPO & \href{https://huggingface.co/datasets/Magpie-Align/Magpie-Llama-3.1-Pro-DPO-100K-v0.1}{Magpie-Align/Magpie-Llama-3.1-Pro-DPO-100K-v0.1}  \\
        HelpSteer2  & \href{https://huggingface.co/datasets/nvidia/HelpSteer2}{nvidia/HelpSteer2}  \\
        OpenMathInstruct-2 & \href{https://huggingface.co/datasets/nvidia/OpenMathInstruct-2}{nvidia/OpenMathInstruct-2}  \\
        LeetCode & \href{https://huggingface.co/datasets/greengerong/leetcode}{greengerong/leetcode}  \\
        Self-Oss-Instruct-SC2 & \href{https://huggingface.co/datasets/bigcode/self-oss-instruct-sc2-exec-filter-50k}{bigcode/self-oss-instruct-sc2-exec-filter-50k}  \\
        Alpaca-GPT4-Zh & \href{https://huggingface.co/datasets/llamafactory/alpaca_gpt4_zh}{llamafactory/alpaca\_gpt4\_zh}  \\
        Magpie-Qwen2-Pro-Zh & \href{https://huggingface.co/datasets/Magpie-Align/Magpie-Qwen2-Pro-200K-Chinese}{Magpie-Align/Magpie-Qwen2-Pro-200K-Chinese}  \\
        \bottomrule        
        \end{tabular}
    }
\label{tab:huggingface_models}
\end{table}

\section{Details of Evaluation Benchmarks}
\label{appendix:benchhmark_details}

\textbf{AlpacaEval-2}~\citep{AlpacaEval} comprises 805 instructions from five different datasets and assesses models using two metrics: length-controlled (LC) win rate and raw win rate (WR)~\citep{dubois2024length}. GPT-4-Preview-1106 serves as both the baseline model and the evaluator for the other models.

\textbf{Arena-Hard}~\citep{arenahard2024} is a challenging instruction-following benchmark that closely aligns with the human preference ranking from Chatbot Arena~\citep{chiang2024chatbot}, a crowd-sourced platform for evaluating LLMs. It spans 250 high-quality topic clusters including 500 well-defined technical problem-solving queries. We report the win rate against GPT-4-0314 using GPT-4-Preview-1106 as the judge model. 

\textbf{MT-Bench}~\citep{zheng2023judging} contains 80 multi-turn dialogues across eight categories, including writing, roleplay, reasoning, math, coding, extraction, STEM, and humanities. Each response is evaluated by GPT-4 on a scale from 1 to 10, with the average score reported for each dialogue turn across the 80 dialogues. We use GPT-4-0613 as the judge model following the official setting.

\textbf{AlignBench v1.1}~\citep{liu2023alignbench} is a comprehensive multi-dimensional benchmark for evaluating LLMs’ alignment in Chinese. It contains 683 high-quality samples spanning 8 main categories, namely fundamental language ability, advanced Chinese understanding, open-ended questions, writing ability, logical reasoning, mathematics, task-oriented role play, and professional knowledge. We employ GPT-4-0613 to analyze and subsequently grade the responses.

\textbf{LiveBench-0831}~\citep{white2025livebench} mitigates test set contamination through monthly question updates and the use of recently released data sources. It features verifiable, objective ground-truth answers for automatic and accurate scoring, eliminating the need for LLM-based evaluation.

\textbf{MMLU-Pro}~\citep{wang2024mmlupro} is an enhanced version of the MMLU~\citep{hendrycks2021mmlu} dataset, designed to address issues such as noisy data and reduced difficulty due to advances in model capabilities and increased data contamination. MMLU-Pro increases challenge levels by expanding multiple-choice options from 4 to 10, requiring reasoning across more questions, and incorporating expert-reviewed annotations for improved quality and reduced noise.

\textbf{MMLU-redux}~\citep{gema2024we} is a re-annotated subset of the MMLU~\citep{hendrycks2021mmlu} dataset created through manual assessment from 14 human experts. 

\textbf{GPQA-Diamond}~\citep{rein2023gpqa} is a challenging knowledge benchmark crafted by PhD-level domain experts in biology, physics, and chemistry. The dataset contains questions that are straightforward for experts but difficult for laypersons. We evaluate the highest quality diamond set comprising 198 questions.

\textbf{GSM8K}~\citep{cobbe2021gsm8k} is a set of grade-school math word questions that evaluates mathematical reasoning capabilities.

\textbf{MATH}~\citep{hendrycks2021math} is a dataset of math problems ranging in difficulty from middle school to high school competition level. It tests a wide range of mathematical skills, including algebra, calculus, number theory, and probability.

\textbf{AMC 23}\footnote{\url{https://huggingface.co/datasets/AI-MO/aimo-validation-amc}}~\citep{yang2024qwen25math} refers to the 2023 American Mathematics Competition, featuring 25 multiple-choice questions that test advanced high school mathematics, including trigonometry, advanced algebra, and elements of calculus.

\textbf{HumanEval}~\citep{chen2021evaluating} evaluates code generation capabilities by presenting models with function signatures and docstrings and requiring them to implement the function body in Python.

\textbf{MBPP}~\citep{austin2021program} is a dataset of simple programming problems designed to assess the ability of models to generate short Python code snippets from natural language descriptions.

\textbf{LiveCodeBench 2408-2411}~\citep{jain2024livecodebench} is a benchmark designed to evaluate coding capabilities using an evolving set of contamination-free problems sourced from platforms including LeetCode\footnote{\url{https://leetcode.com}}, AtCoder\footnote{\url{https://atcoder.jp}}, and CodeForces\footnote{\url{https://codeforces.com}}. We evaluate the subset comprising 160 problems published between August 2024 and November 2024.

\end{document}